\pdfoutput=1 
\documentclass[aps,amsmath,amssymb,preprintnumbers,preprint,nofootinbib,a4paper,11pt]{article}
\usepackage{jheppub} 

\usepackage[T1]{fontenc} 

\usepackage{float}
\usepackage{caption}
\usepackage{subcaption}
\usepackage{multirow}
\usepackage{booktabs}
\usepackage{multicol}
\usepackage{times}
\usepackage{epsfig}
\usepackage{graphicx}
\usepackage{amsmath}
\usepackage{amssymb}
\usepackage{wrapfig}
\usepackage[T1]{fontenc}
\usepackage[font=small,labelfont=bf,tableposition=top]{caption}
\DeclareCaptionLabelFormat{andtable}{#1~#2  \&  \tablename~\thetable}
\newcommand{\fastai}{{\bf \color{blue} Fast.AI }} 
\title{\boldmath Developing an App to interpret Chest X-rays to support the diagnosis of respiratory pathology with Artificial Intelligence}

\author[1]{Andrew Elkins }
\author[2]{Felipe F. Freitas }
\author[3]{Ver\'onica Sanz}
\affiliation[1]{Brighton and Sussex University Hospitals, NHS Trust, Brighton
BN2 5BE, United Kingdom}
\affiliation[2]{ CAS Key Laboratory of Theoretical Physics, Institute of Theoretical Physics, Chinese Academy of Sciences, Beijing 100190, China}
\affiliation[3]{Alan Turing Institute,  British Library, 96 Euston Road, London NW1 2DB \& \\Department of Physics and Astronomy, University of Sussex, Brighton BN1 9QH, United Kingdom}

\emailAdd{Andrew.Elkins@bsuh.nhs.uk}
\emailAdd{fp4303@itp.ac.cn}
\emailAdd{v.sanz@sussex.ac.uk}
\abstract{ In this paper we present our work to improve access to diagnosis in remote areas where good quality medical services may be lacking. We develop new Machine Learning methodologies for deployment onto mobile devices to help the early diagnosis of a number of life-threatening conditions using X-ray images.  By using the latest developments in fast and portable Artificial Intelligence environments, we develop a smartphone app using an Artificial Neural Network to assist physicians in their diagnostic.}

\begin{document} 
\maketitle
\flushbottom

\section*{Introduction}

Medical images, including results from X-rays, are an integral part of medical diagnosis. Their interpretation requires an experienced radiologist, a human whose skills are scarce and who can commit mistakes when tired. But at face value, the X-ray diagnostic uses simple features from the image, such as black and white intensity, contours and shapes, all properties which  an Artificial Intelligence (AI) can handle well. 

And indeed, in recent years, the idea of using AI as a means of assisting diagnostic (classification) has taken hold, thanks to the increase of computational speed (e.g. with GPUs), the availability of large and well-documented datasets, and the development of deep learning techniques, in particular Convolutional Neural Networks (CNN). The results of a number of studies training CNNs to diagnose diseases using 2D and 3D images are rather impressive, reaching near 100\% accuracy in such diverse pathologies as lung nodules or Alzheimers, see Ref.~\cite{review} for a recent overview. Moreover, in some cases the performance of the AI exceeds that of radiologists, see e.g. the study of CheXNet~\cite{chexnet}.

 One of the main problems faced by people in developing countries is access to timely medical diagnosis. Lack of investment in health care infrastructure, geographical isolation and shortage of trained specialists are common obstacles to providing adequate health care in many areas of the world. 
 
 Yet the use of AI is still limited for various reasons, technical and sociological. For example, images are only one aspect of the patient history which can be supplemented with clinical signs, e.g. whether the patient has fever. Currently, the availability of databases with additional clinical information is scarce, but databases with better labelling could substantially improve the AI training in the near future. A more important issue is related to the natural variability of a large training dataset, and the need to remove modelling when interpreting the image. This has led to the development of new techniques beyond CNNs (supervised learning) to incorporate a more bottom-up approach: unsupervised machine learning. Nevertheless, all these studies rely on a large number of neuron layers, typically dozens, and good quality images. The trained neural network  can then be used in a powerful computer station to help diagnosis.
 
 Given our interest in portability, we need to develop a different strategy. We cannot deploy machine learning algorithms which depend on too many layers, and we cannot rely on the acquisition of precise images. Instead, we have to strike a balance between the desired outcomes (portability and reliability) and the real-life situations a clinician may encounter on the field.
 
 This note presents our results to help the early diagnosis of a number of life-threatening conditions in remote areas with the use of new methodologies (based on Machine Learning) and their capability to be deployed into mobile devices.  We use the latest developments in AI environments which are portable and fast, in particular \fastai~\cite{fastai}, to develop an Artificial Neural Network  to be deployed as a smartphone app, or in one of Google's AI development kits, or in Rasberry PI, as a tool to assist physicians in their diagnostic. A beta version of the app can be tested in the {\tt heroku} environment~\cite{heroku} and all the code developed during this project can be obtained in {\tt GitHub}~\cite{github}.

\section{Analysis set-up}

\subsection{The dataset and analysis framework}
We use the ChestX-ray14 dataset~\cite{Wang} which contains 112,120 frontal-view X-ray images of 30,805 unique patients. The images are frontal view of chest X-rays PNG images in 1024$\times$1024 resolution. Meta-data for all images contains: Image Index, Finding Labels, Patient ID, Patient Age, Patient Gender, View Position, Original Image Size and Original Image Pixel Spacing. Each image contains the annotations up to 14 different thoracic pathology labels. The labels were assigned using automatic extraction methods on radiology reports.The database from NIH is available online through the link: \url{https://nihcc.app.box.com/v/ChestXray-NIHCC/folder/36938765345}. 

The 14 common thorax disease categories are: Atelectasis,  Cardiomegaly, Effusion,  Infiltration, Mass,  Nodule,  Pneumonia, Pneumothorax,  Consolidation, Oedema,  Emphysema,  Fibrosis,  Pleural Thickening, and Hernia.

Regarding the analysis of the dataset, in this project we use AI techniques in the framework of \fastai~\cite{fastai},  a library which simplifies accurate and fast training of NNs using modern best practices. It includes out-of-the-box support for vision, text, tabular, and collaborative filtering models. Another main advantage of \fastai is the simplification of the data processing made by the \verb|datablock| API, and the implementation of the fit one cycle method~\cite{Leslie} and mixed precision training~\cite{Micikevicius}. The  fit one cycle method allows us to drastically reduce the training time by allowing changes during the training in the learning rates and momentum (hyper-parameters). This method can reduce the time to train big NN models, but also stabilise it by helping the optimisation algorithm to prevent falling into saddle points. The  mixed precision training uses half-precision floating point numbers, without losing model accuracy or having  to modify hyper-parameters. This nearly halves memory requirements and, on recent GPUs, speeds up arithmetic calculations.

\subsection{Selecting and processing the data}

Since we are interested in distinguishing different types of diseases, we will focus on the dataset with labels of at least one disease. In the dataset there are 51759 images with at least one disease (non- healthy). On this dataset, we will perform two types of analyses: {\it 1.) } one vs all and {\it 2.)} multi-label classification. 

In the first type of analysis, we will train an algorithm to identify patients with a specific disease, which we choose to be {\tt Pneumothorax} for illustrative purposes, versus any other type of disease in the sample.  
In the second type of analysis, we will tackle a more difficult but realistic situation: often diseases do not occur in isolation, e.g.  {\tt Atelectasis} tends to appear in conjunction with {\tt Cardiomegaly} or {\tt Consolidation}, or both. This more complex situation is called {\it co-occurrence}, which in terms of the data analysis corresponds to a problem of multi-label classification.

Below we describe how we select and process the data in these two situations:

\begin{center}
\subsubsection{ One vs All problem [i.e. {\tt Pneumothorax} vs All]}
\end{center}

We separate the dataset in two categories, one with the label {\tt Pneumothorax} the other with all the images with no {\tt Pneumothorax} label ({\tt Atelectasis, Pneumonia, ...}).  Specifically, we first read from the metadata information available in the file \verb|Data_Entry_2017.csv| the images names and findings, selecting the entries with the {\tt Pneumothorax } label to one folder and all the others to a different folder. After this selection one ends up with the following set of samples:
\begin{itemize}
\item Pneumothorax: 5302 images, resolution 1024x1024 8-bit gray scale
\item No-Pneumothorax: 46457 images, resolution 1024x1024 8-bit gray scale
\end{itemize}

Next, we prepare the data for testing and validation using  the \fastai  \verb|DataBlock| API, which automatically loads the data from their respective folders, assign their labels according to the name of the folders and split the full data into training (80\%) and validation (20\%) datasets. The \verb|DataBlock| API ensure the use of correct labels, the correct train/validation (or test) split, the correct normalisation of the images and all the augmentations transformation we later apply~\footnote{Note that the pixel values can vary from 0 to 255 for each of the three channels in a RGB image, or one single channel for Gray-Scale, and we re-scale this range to -1 to 1 to input the values in a CNN.}.
After this selection and transformations we end up with:
\begin{itemize}
\item Train size (80\%): 41408 items, items dimension: 224x224 3 channels, transformation applied.
\item Validation size (20\%): 10351 items, items dimension: 224x224 3 channels, no transformations.
\item Batches which are set to 64 items per batch, so they can fit on a regular GPU memory.
\end{itemize}

We select the following transformations to be applied to the train dataset:
\begin{itemize}
\item random rotation (clockwise or counter-clockwise) in an angle range between 0 and 30 degrees.
\item 50\% chance of an image to be zoomed by a 1.3 scale.
\item 100\% chance of a image be selected to randomly modify brightness and contrast within a range between 0 and 0.4.
\item normalize the items according to the stats of each batch, i.e. take the mean and the standard deviation of each channel and normalize the image using them\footnote{Note that in pytorch when we normalize images we have to provide the mean and the std for each channel, in tensorflow we can use  {\tt sklearn.preprocessing.StandardScaler} to do the same.}.
\end{itemize}

Note that the images fed into the CNN are set to be $224\times 224$ with three channels due to hardware limitations and the size of the batch (64 images) to fit into the GPU memory, as well as coding implementation -- the CNN architecture we are using (DenseNet) was originally designed to work with 3 channels (RGB), so the input layer is designed to take inputs (tensors) with the dimension (batch size,C=3,x size, y size) where C is the number of channels from the image. 

\begin{figure}[t!]
\centering
\includegraphics[width=.7 \linewidth]{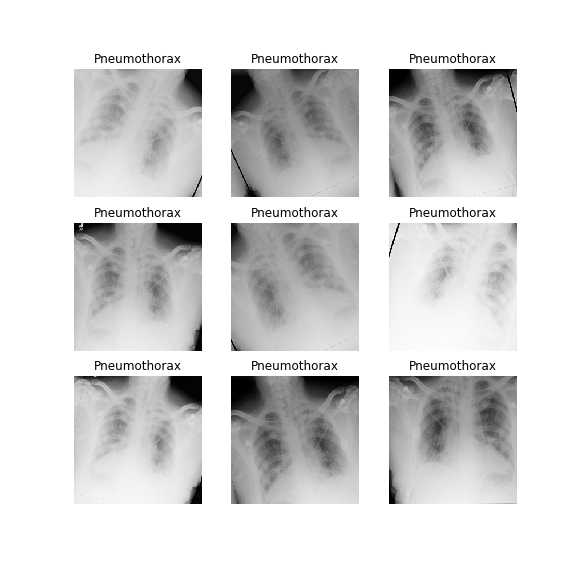}
\caption{Random transformations applied in one selected image. The random rotation and change in brightness are applied to consider the variations of image quality and arrangement one can encounter in medical facilities in remote areas.}
\label{fig1}
\end{figure}

In Fig.~\ref{fig1} we show the effects of the transformations we apply in the train dataset items. They are designed to emulate some of the issues which different situations could lead to. For example, brightness variance can occur when the technician exposes the patient to a more energetic X-ray beam (higher keV X-rays) or long exposure time, which greatly increases the contrast of the image, but also increases the radiation dose received by the patient; or in the case of film X-ray, the exposure time to chemicals can drastically reduce contrast of the image, leading to erroneous feature differentiability.

\begin{center}
\subsubsection{Multi-label classification}
\end{center}

For the multi-label case we prepare a new csv file which contains the name of the image (in our input dataset the names are formatted as \verb|00025252_045.png|) and the label of the diseases found in the respective images. This file will be used by the \verb|DataBlock| API to correctly label the images. The file \verb|Data_Entry_2017.csv| contains all the metadata with the name of each image together with the findings and other information that can be used for future applications. 

To create the multi-class label file we use a python script with two main functions: one to extract the information from the \verb|Data_Entry_2017.csv| metadata and another function to write the new csv file with the columns contain the path to the images, their respective findings and a new column with a one-hot encode\footnote{In digital circuits and machine learning, one-hot is a group of bits among which the legal combinations of values are only those with a single high (1) bit and all the others low (0).} to represent the presence or not of each disease (class) in a given image. All the scripts will be available in {\tt GitHub} so anyone can reproduce our findings. 

\begin{figure}[h]
\centering
\includegraphics[width=1\linewidth]{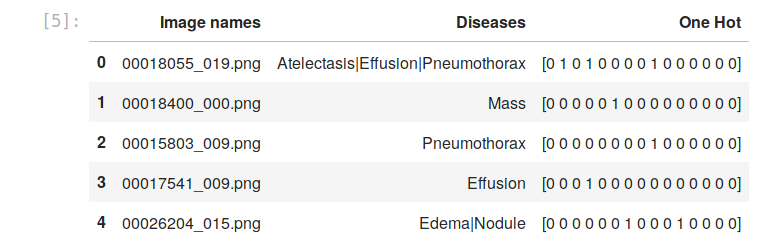}
\caption{Extract of the label dataframe for the multi-label classification task. \fastai can handle multi-label classification targets and image names are given in a proper structured way, which we provide a new csv file. The one-hot encode represents the presence (1) or absence (0) of a particular disease in the list: Atelectasis, Cardiomegaly, Effusion, Infiltration, Mass, Nodule, Pneumonia, Pneumothorax, Consolidation, Edema, Emphysema, Fibrosis, Pleural Thickening and Hernia.}
\label{fig1a}
\end{figure}

The \fastai API contains a very convenient method to prepare and load the images into our CNN model called \verb|ImageDataBunch|, which and contains a series of useful functions which one can use for AI purposes. For example, for the multi-label classification one can use the \verb|ImageDataBunch.from_csv()|, which automatically identifies  the images and  multi-class labels we are loading into the CNN, e.g.

\begin{verbatim}
np.random.seed(42)
data = ImageDataBunch.from_csv(path, label_delim='|', 
           csv_labels='label list.csv', label_col=1,\
           delimiter=',', valid_pct=0.2, ds_tfms=tfms, 
           bs=64, size=224, num_workers=3).normalize()
\end{verbatim}

This set of commands will automatically load the images in batches of 64, re-size the images form 1024$\times$1024 to 224$\times$224, load the labels from \verb|label list.csv| and normalise according to the {\tt pytorch} standard, and split the dataset into 80\% test images and 20\% for validation.

\begin{figure}[h]
\centering
\includegraphics[width=.75\linewidth]{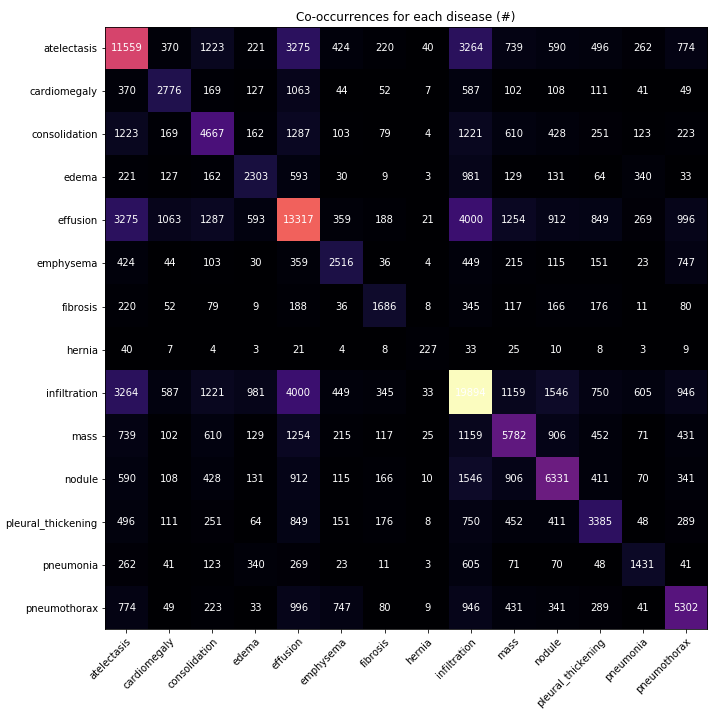}
\caption{Number of co-occurrences for each disease in the dataset. Note, for example, that Infiltration is always co-occurrent with some other disease label, and that some diseases often come together, e.g. Effusion and Atelectasis.}
\label{fig1b}
\end{figure}

A very important  aspect in our analysis is how often one disease appears together with others, i.e. the rate of {\bf \it co-occurrences}. We can check the co-occurrences of a disease using {\tt scikit-learning} feature extraction library, and in Fig.~\ref{fig1b} we show the matrix of number of co-occurrences in the dataset. This information is key to introduce the proper weights used in the loss function and lead to the positive identification of a given class. These values can help us to better calibrate the weights for each class in order to avoid situations like a  model who can only predict the class where one has the higher number of targets.

\subsection{The model architecture and training}
\begin{figure}[h]
\centering
\includegraphics[width=1\linewidth]{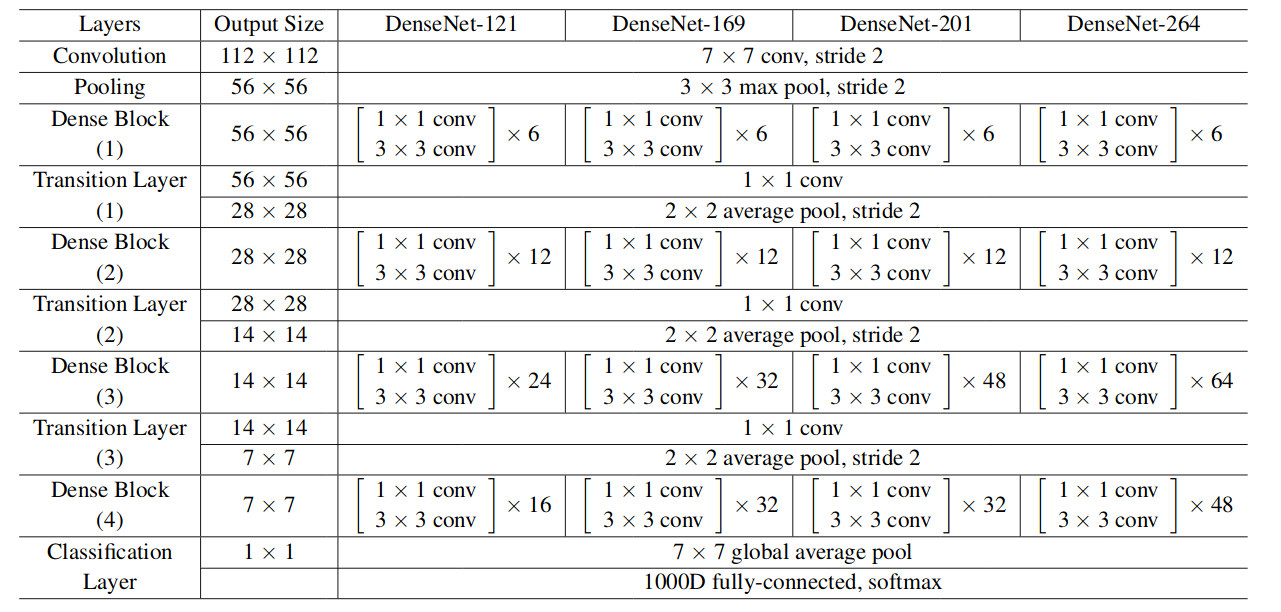} 
\caption{DenseNet architectures for ImageNet. Note that each ``conv'' layer shown in the table corresponds the sequence: Batch Normalization, Rectifier Linear function and a Convolutional Layer (BN-ReLU-Conv). }
\label{fig2}
\end{figure}

Our architecture will be based on the DenseNet-121 model with the last layers (header) modified for our purposes. Specifically, our convolutional neural network (CNN) model consists on the following parts:
\begin{itemize}
\item The body of the model is the same as the denseNet-121, described in table~\ref{fig2}  (third column) (DenseNet-121).
\item For the classification layer, we remove the last layer from denseNet-121 and replace it with the following layers:
\begin{itemize}
\item  AdaptiveConcatPool2d layer
\item  Flatten layer
\item  A block with [nn.BatchNorm1d, nn.Dropout, nn.Linear, nn.ReLU] layers.
\end{itemize}
\end{itemize}

In our loss function we use the weights from each class calculated according to the {\tt scikit-learn} class weight method, defined as:
\begin{eqnarray}
\frac{n_{samples}}{(n_{classes} * [n_{1},n_{2},...])}  
\label{weightsC}
\end{eqnarray}
Where $n_{samples}$ is the total number of images in the training sample, $n_{classes}$ is the number of classes (2 in the case of One vs All), and $n_{1},n_{2},...$ are the number of images for each class. This function from {\tt sklearn} returns the weight for each class in our training sample, information we pass onto the weights in the loss function, which mitigates the problem of having more images from one class than another class, i.e. imbalanced classes datasets.

\begin{figure}[h]
\centering
\includegraphics[width=1\linewidth]{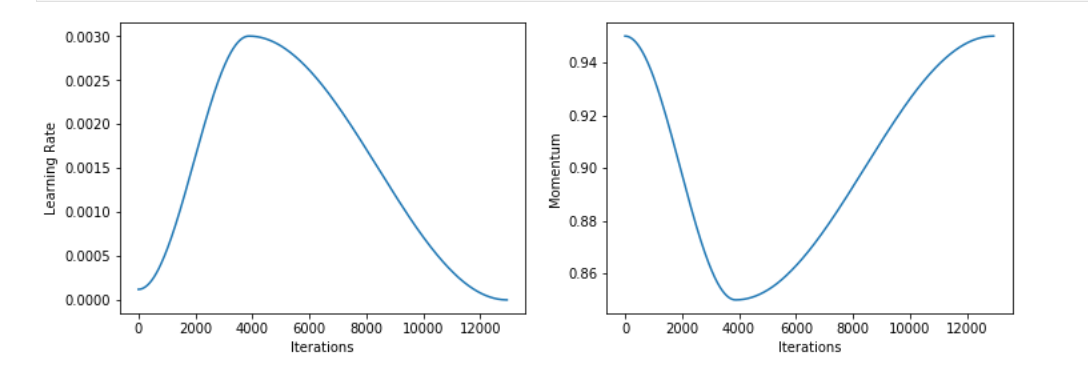}
\caption{{\it Fit one cycle method}. Left figure: this shows the variation of  the learning rate over the number of iterations, with the vertical axis showing the learning rate change between $1e^{-4}$ to $3e^{-3}$ and the horizontal axis showing all the iterations (i.e. epochs). Right figure:  here we show the change in momentum rate over number of iterations. In these figures we can see the change in values of the learning rate and momentum,  during the middle of the cycle. The high learning rates and small momentum will act as regularisation method, and keep the network from overfitting, as they prevent the model from landing in a steep area of the loss function, preferring to find a minimum that is flatter.}
\label{fig3}
\end{figure}

The network is trained end-to-end using Adam as an optimizer \cite{Diederik} with standard parameters ($\beta_{1}$ = 0.9 and $\beta_{1}$ = 0.999). We trained the model using the fit one cycle method \cite{Leslie}, which consist in the following steps: initialises the training with a given learning rate and momentum; in the middle of the training phase the learning rate is increased up to a given maximum value while the momentum is reduced; and close to the end of the training the learning rate is reduced again and the momentum increased, as shown in figure \ref{fig3}. This method gives more stable results by avoiding the optimization process to fall into saddle points and requires less epochs to fully train our model.

Another useful parameter for training our CNN is the weight decays\footnote{Do not confuse this term with class weights.  The class weights are use to balance the number of images of each class.}. After each epoch, the weights of the NN are multiplied by a smaller factor between 0 and 1. This is one of the various forms of regularisation of the NN training, together with batch size and dropout\cite{Nitish}. One can choose a weight decay which allows us to use a bigger initial value for the learning rate and thus reduce the time of training during the fit one cycle.

For the One vs All case we choose the Cross Entropy Loss with weights defined for each class to take into account the high imbalance between the Pneumothorax case vs all the others. Specifically, to apply the weights into our loss function we do as follows:
\begin{verbatim}
class_weights = torch.FloatTensor(weights).cuda()
loss_func = nn.CrossEntropyLoss(weight=class_weights)
\end{verbatim}
where the weights are calculated by the equation \ref{weightsC}, the function \verb|cuda()| just moves the numbers calculated for the weights to the GPU memory where our CNN is loaded. The weights are passed to the loss function by the argument \textbf{weight=} from the cross entropy loss class.

\begin{figure}[h]
\centering
\includegraphics[width=0.45\linewidth]{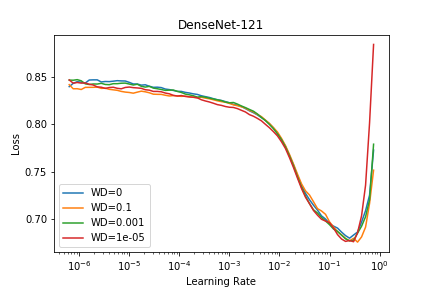}
\includegraphics[width=0.45\linewidth]{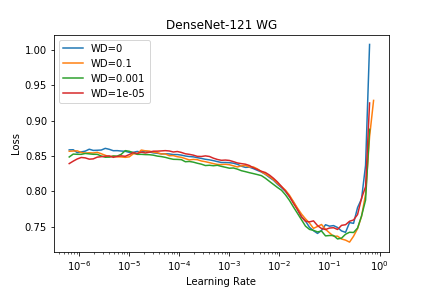}
\caption{Effect of the different choices for the weight decay (WD) in the loss change (vertical axis) for a given learnnig rate (horizontal axis) for the DenseNet-121 CNN without class weights (left panel) and with class weights (right panel). Note that the loss is relatively insensitive to the choice of WD, which is a sign of stability of our model. With small values for the WD one can choose large values for the learning rate and reduce the training time (i.e. number of epochs).}
\label{fig2a}
\end{figure}

The Cross Entropy Loss function, in the case of the class weight argument being specified, can be described as:
\begin{eqnarray}
\text{loss}(x, class) = weight[class] \left(-x[class] + \log\left(\sum_j \exp(x[j])\right)\right)
\end{eqnarray}
with $x$ as the image input for the respective class.

Another important aspect for the Cross Entropy Loss is the reduction method, which specifies the reduction to apply to the output. The {\tt pyTorch} framework gives three options: $none,mean$ and $sum$, and we used $mean$ in our analysis. For the multi-label classification task we choose the Binary Cros Entropy loss with  \verb|BCEWithLogitsLoss|. This loss combines a Sigmoid layer and the \verb|BCELoss| in one single layer. This loss is more stable numerically than a plain Sigmoid layer followed by a \verb|BCELoss|. By combining the operations into one layer, one takes advantage of the log-sum-exp trick \cite{Nielsen} for numerical stability.

The \verb|BCEWithLogitsLoss| for the multi-label case with class weights is described by:
\begin{eqnarray}
        &\ell_c(x, y) = L_c = \{l_{1,c},\dots,l_{N,c}\}^\top,\nonumber \\ 
        &l_{n,c} = - w_{n,c} \left[ p_c y_{n,c} \cdot \log \sigma(x_{n,c})
        + (1 - y_{n,c}) \cdot \log (1 - \sigma(x_{n,c})) \right] \ ,
\end{eqnarray}
where $c$ is the class number ($c > 1$ for multi-label binary classification, $c = 1$ for single-label binary classification), $n$ is the number of the sample in the batch and $p_c$ is the weight of the positive answer for the class $c$. Note that $p_c > 1$ increases the recall, whereas the choices with $p_c < 1$ increase the precision. Also note that we used the default option $mean$ as reduction method to \verb|BCEWithLogitsLoss|.

Before training, one has  to define the range of the learning rate to be used in the fit one cycle method. To  compute the best range for the learning rate we used the function \verb|learner.lr_find()|, which will perform a test run, starting with a very small learning rate and increasing it after each mini-batch until the loss function starts exploding. Once the loss starts diverging, the \verb|lr_find()| will stop the range test run. In figure \ref{fig2a} we show the loss values versus the learning rate for each weight decay choice (WD$=0,0.1,0.001, 1e^{-5}$). The best initial values for learning rates are the ones giving the steeper gradient towards the minimum loss value \cite{Leslie}. In the case of figure \ref{fig2a} this value is $1.32e^{-2}$ for the left figure and $1.02e^{-2}$ for the right one.

The learning rate change can be described as follows:
\begin{eqnarray}
&n = \text{number of iterations} \nonumber \\
&\verb|max_lr| = \text{maximum learning rate} \nonumber \\
&\verb|init_lr| = \text{lower learning rate (we start the range test from this value)} \nonumber \\
&\verb|max_lr| = \verb|init_lr|*q^{n}\nonumber \\
&q = (\verb|max_lr|/\verb|init_lr|)^{\frac{1}{n}}
\end{eqnarray}

Note that the learning rate after the i-th mini-batch is given by:
\begin{equation}
lr_{i}= \verb|init_lr|*(\verb|max_lr|/\verb|init_lr|)^{\frac{i}{n}}
\end{equation}


We are using the transfer learning methodology to not just training our CNN faster, but also to increase the accuracy. To execute this task we trained the CNN in two phases. During the first phase, the model is trained from end-to-end, i.e. all the layers are trained, for 30 epochs with a initial learning rate of 1.32e-02 to a maximum 1e-1. This initial value is chosen due to the loss value for this learning rate exhibit the steeper gradient value towards the direction of the minimum, as we can see in figure \ref{fig2a}. After the 30 epochs, the training is stopped and the weights are saved, leading to a second train phase which uses the weights saved from the previous training, freeze all the layers before the last classification layer to void retraining all the NN, and train only the last classification layers with a different range for the learning rate. This process described above is what we know as {\tt transfer learning}, and  we evaluate the impact of this methodology on the accuracy of our model between the two phases.


\section{Results}

In a simple binary classification problem (like in our benchmark Pneumothorax vs All), one can define some measures of how well our algorithm is performing: precision, accuracy and recall. All these are computed by evaluating how often in a sample the results lead to a true-positive ({\bf TP}), true-negative ({\bf TN}), false-positive ({\bf FP}) and false-negative ({\bf FP}) diagnoses. Based on these numbers one can define three measures of goodness of diagnosis:
  
 {\bf Accuracy = TP+TN/TP+FP+FN+TN}, the ratio of correct observations versus all observations, e.g. accuracy would tell us how likely is to correctly identify pneumothorax pathologies in sick patients.
  
{\bf Precision = TP/TP+FP}, the ratio of correct predicted positives versus all the classified as positive, e.g. precision would tell us how many patients diagnosed with pneumothorax do actually have pneumothorax.

{\bf Recall (or sensitivity) = TP/TP+FN}, the ratio of correctly predicted positive versus all the elements with the actual disease, e.g. recall would tell us the probability for correctly diagnosing pneumothorax among all the patients suffering from pneumothorax.

Obviously all these measures are important, and depending on the nature of the disease and focus of the practitioner the algorithm could be trained to improve any of these three, usually at the cost of the other two. A typical {\it average} measure of goodness which works well in imbalanced datasets is the F1 score, defined in its simplest form as an average between accuracy and recall 
 \begin{center}
 {\bf 
F1 score= 2 * (precision * recall) / (precision + recall) }
 \end{center}

\subsection{Results for One vs All (Pneumothorax)}

We can compute the F1 scores and Precision-Recall of our CNN using our validation dataset. In scikit-learn, F1 scores can be evaluated using different average methods\footnote{\url{https://scikit-learn.org/stable/modules/generated/sklearn.metrics.f1_score.html}}, namely:
\begin{itemize}
\item[binary:] Only reports results for the class specified by positive label, i.e. Pneumothorax detected. This method is only applicable in the binary problem (Pneumothorax vs All) but not in the multi-class problem. 
\item[micro:] This method counts the total true positives, false negatives and false positives, to compute a more global metric.
\item[macro:] In this case, the function calculates the metrics for each label and find their unweighted mean, i.e. it does not take label imbalance into account.
\item[weighted:] In this case, the function finds metrics for each label, and their average weighted by the number of true instances for each label. This is an improvement from 'macro' to account for label imbalance.
\item[samples:] In this case, the function calculates metrics for each instance, and find their average. This is only meaningful for multi-label cases. 
\end{itemize}

Since we are dealing with high imbalanced classes we will be  using the weights for our samples from the validation dataset:
\begin{equation}
W_{\text{validation}} = [\text{All others}: 0.5566, \text{Pneumothorax}:4.902]
\end{equation}
with these weights we can build an array with the same size as the ground labels and evaluate all the metrics  taking into account the class imbalance. The results are as follows 
\begin{itemize}
\item F1 score (macro): 0.5822911828419564
\item F1 score (micro): 0.6762802817903739
\item F1 score (weighted): 0.7132615220206845 
\item F1 score (binary): 0.384149629028795
\item F1 score (All others, Pneumothorax): [0.780433 0.38415]
\end{itemize}
and show the range of different F1 scores one can obtain, even in the binary classification problem.

\begin{figure}[h]
\centering
\includegraphics[width=0.55\linewidth]{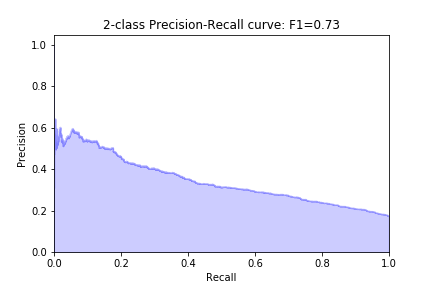}
\includegraphics[width=0.55\linewidth]{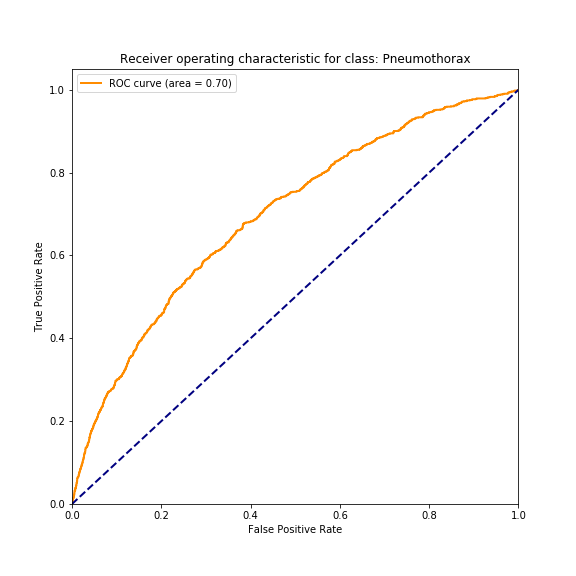}
\caption{Precision-Recall ({\it top}) and ROC ({\it bottom}) curves for our model in the Pneumothorax vs All case.}
\label{fig7}
\end{figure}

There are two common representations of the goodness of the algorithm. One is the  precision-recall curve, shown in Fig.~\ref{fig7}. A high area under the curve represents both high recall and high precision and is an indication of good performance.

 Another representation is the so-called ROC curve (Receiving Operating Characteristic) which shows the shape of TP as a function of FP. One often says that the algorithm {\it learns} when the curve is steeper than a straight line, namely the algorithm is doing better than a 50\% chance of identifying the right disease (better than a random pick). In Fig.~\ref{fig7}, the orange line corresponds to the ROC curve of our algorithm, which is clearly performing better than chance. A related and more global measure of goodness based on the ROC plot is the AUC (Area Under the Curve), and in this figure AUC is 70\% for detecting Pneumothorax.

\subsubsection{Layers heatmaps}
To visually understand what regions in the image our CNN is picking up on, we look at the area of the image which is  more active when we show an image from a given class. To do this, we built a Grad-CAM following Ref~\cite{gradcam}.

\begin{figure}[h]
\centering
\includegraphics[width=0.45\linewidth]{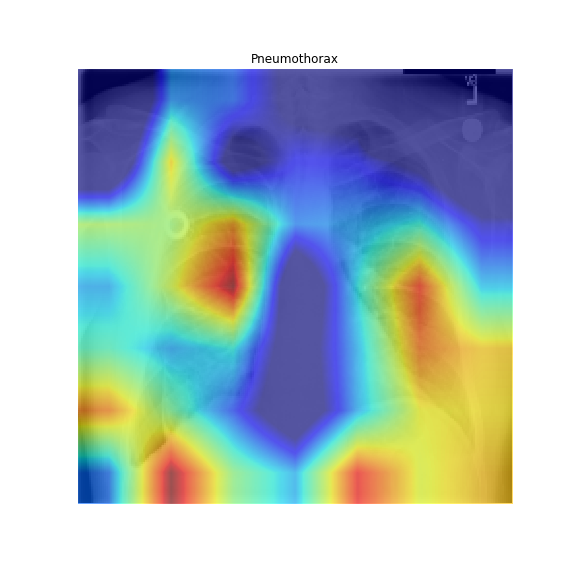}
\includegraphics[width=0.45\linewidth]{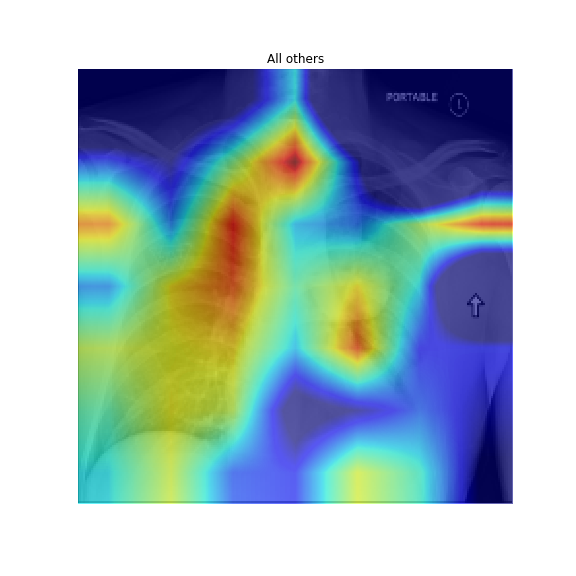}
\caption{Grad-CAM for a Pneumothorax item (left) and no Pneumothorax item (right) class. The regions in red show the areas which activate more units (neurons) in the last convolutional layer before the classification.}
\label{fig8}
\end{figure}

The regions displayed in yellow to red colours highlight the spatial location of the features which activate more intensely the last convolutional layer before the classification. As one can see, in the Pneumothorax image the CNN is identifying important features located in the left lung, top of diaphragm and a region below the right lung. Meanwhile, for the non-Pneumthorax image the CNN identify as {\it hot regions} on the top of diaphragm, apex of the heart and a hot spot on the left side outside of the body. These images provide valuable information about what are the most important features for our CNN and how to tune the parameters of the model to improve the training phase. One conclusion we can extract from these images is that, although overall we are getting good classification scores, the CNN is struggling to localise good features to better classify the images. We plan to use this method to   improve on the training, by including more random transformation like flip horizontal and vertical, random crop and a different pad method.

\subsection{Results for multi-label}

We move onto a much more complex case, where instead of a binary classification problem Pneumothorax vs no-Pneumothorax, we tackle the classification of 14 diseases, including the additional difficulty of co-occurrence of various diseases in the same patient. 

The first thing we need to do is to define new measures of goodness for our method. Precision-Recall or ROC curves like Fig.~\ref{fig7} are designed to understand a binary classification problem. 

We need to generalise these two notions to a multi-classification case. For example, one ROC curve can be drawn per label, but one can also draw a ROC curve by considering each element of the label indicator matrix as a binary prediction (micro-averaging).
Another evaluation measure for multi-label classification is macro-averaging, which gives equal weight to the classification of each label. 

\begin{figure}[h]
\centering
\includegraphics[width=0.6\linewidth]{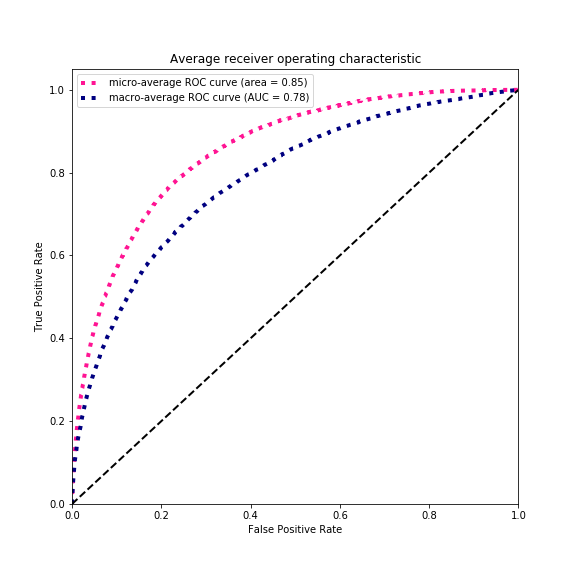}
\caption{Micro and macro ROC average for multi-label class. For the micro-average we got AUC=82\%, while for macro-average we obtain AUC=72\%}
\label{fig9}
\end{figure}

The 'micro-average', purple dashed curve Fig.~\ref{fig9}, is calculated as each element of the label indicator matrix, in our case a matrix with the dimensions (number of validations samples, number of classes), as a binary prediction. For the 'macro-average' we first need to aggregate all elements from the false positive rate (fpr) of each class and casting them into a list which we call \verb|all_fpr|. From this list we can interpolate each aggregated fpr and the ROC curves of each class we have calculated. By averaging the interpolated points computed previously over the number of classes (14 for our data), we obtain the mean of the true positive rate for all classes (blue dashed curve Fig.~\ref{fig9}). 

\begin{figure}[h]
\centering
\includegraphics[width=0.72\linewidth]{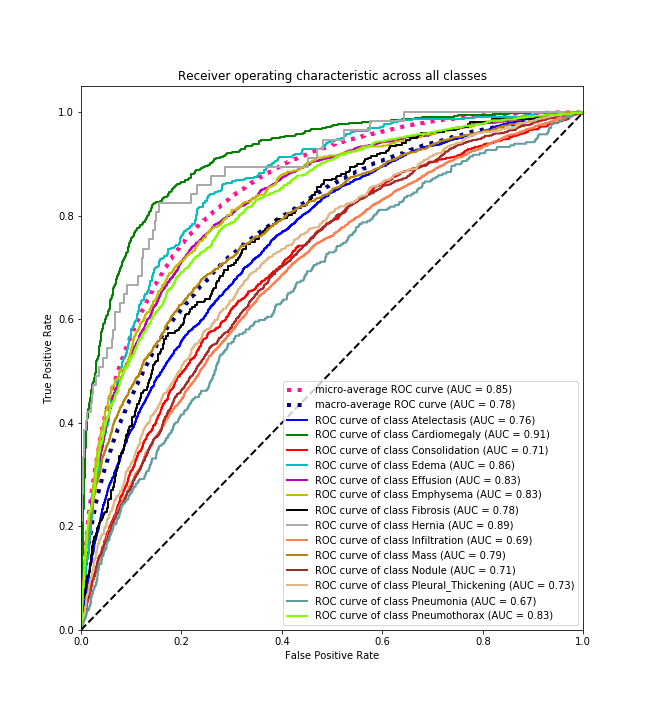}
\caption{ROC curve for each of 14 classes and their respective AUC. The higher AUC we get is for Cardiomegaly AUC=85\%, while the lowest we get is for mass (AUC=56\%), Pneumothorax we got AUC=74\%}
\label{fig10}
\end{figure}

We can also compute the ROC and AUC for each class using the predictions and the truth information for each class. The results are shown in Fig.~\ref{fig10}, where we display all ROC and AUC for the 14 classes including the micro-average and macro-average.

\begin{figure}[h]
\centering
\includegraphics[width=0.6\linewidth]{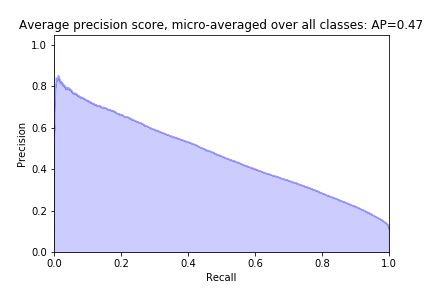}
\caption{Precision-Recall curve considering micro-average, i.e. an aggregate of the contributions from each class and average over all classes. In a multi-class classification, micro-average is preferable to macro-average for class imbalance.}
\label{fig11}
\end{figure}

\begin{figure}[h]
\centering
\includegraphics[width=1.1\linewidth]{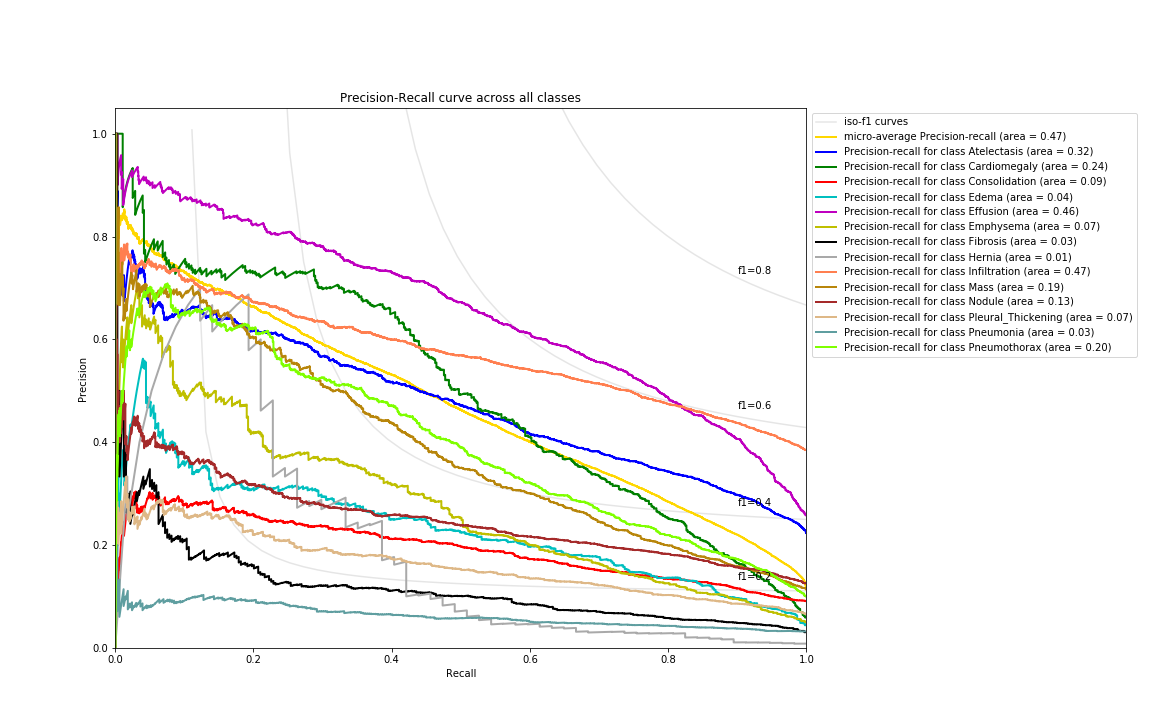}
\caption{Precision-Recall curves for each of the 14 classes and their respective areas. The areas are calculated using the average precision score considering the weights of each class. The iso-curves show the F1 scores in the Precision-Recall plane.}
\label{fig12}
\end{figure}

As discussed in the previous section (Pneumothorax vs All), the Precision-Recall plot is a useful measure of prediction success when the classes are very imbalanced. In Figures \ref{fig11} and \ref{fig12}, we show the precision-recall curves for the average and for each class, respectively.   

The average precision (AP) summarizes the precision-recall plots as the weighted mean of precisions achieved at each threshold, with the increase in recall from the previous threshold used as the weight:
\begin{equation}
\textbf{AP} = \sum_{n}(R_{n} - R_{n-1})P_{n}
\end{equation}
wheree $R_{n}$ and $P_{n}$ are the precision and recall at the n-th threshold.

Finally, in Tables \ref{tab:comparison} and \ref{tab:comparison2} we quote the values of the AUC score and the average precision for each class. 

\begin{table*}[ht!]    
\centering
\begin{tabular}{l c c |}
\toprule
Pathology &  DenseNete-121-\fastai (30-epochs)\\
\midrule
\midrule
 Atelectasis & \textbf{0.76} \\
 Cardiomegaly   & \textbf{0.91} \\
 Effusion  & \textbf{0.83} \\
 Infiltration   & \textbf{0.69} \\
 Mass    & \textbf{0.79} \\
 Nodule     & \textbf{0.71} \\
 Pneumonia     & \textbf{0.67} \\
 Pneumothorax     & \textbf{0.83} \\
 Consolidation    & \textbf{0.71} \\
 Edema    & \textbf{0.86} \\
 Emphysema   & \textbf{0.83} \\
 Fibrosis    & \textbf{0.78} \\
 Pleural Thickening    & \textbf{0.73} \\
 Hernia    & \textbf{0.89} \\
\bottomrule
\end{tabular}
\caption{ \fastai AUC score for each class in the ChestX-ray14 dataset.}
\label{tab:comparison}
\end{table*}

\begin{table*}[ht!]    
\centering
\begin{tabular}{l c}
\toprule
Pathology & DenseNete-121-\fastai (30-epochs)\\
\midrule
\midrule
 Atelectasis  & \textbf{0.31} \\
 Cardiomegaly  & \textbf{0.21} \\
 Effusion   & \textbf{0.42} \\
 Infiltration  & \textbf{0.46} \\
 Mass    & \textbf{0.13} \\
 Nodule    & \textbf{0.13} \\
 Pneumonia    & \textbf{0.03} \\
 Pneumothorax    & \textbf{0.17} \\
 Consolidation   & \textbf{0.09} \\
 Edema    & \textbf{0.04} \\
 Emphysema   & \textbf{0.07} \\
 Fibrosis     & \textbf{0.03} \\
 Pleural Thickening   & \textbf{0.07} \\
 Hernia      & \textbf{0.01} \\
\bottomrule
\end{tabular}
\caption{Average precision for each class.}
\label{tab:comparison2}
\end{table*}

\subsubsection{Layers activation maps and Guided-Grad-CAM}

Here we perform a similar study as in the Pneumothorax vs All case by identifying the regions in the image which activates neurons in the CNN. All the images used in to generate the activations and Grad-CAM maps are from the validation data set, namely these are images the CNN never {\it saw}  before. In the next figures (\ref{fig15}-\ref{fig27}),  we show examples for different pathologies: the input figure, the Grad-CAM heatmap, and the PR curve related to the specific class.

\begin{figure}[h]
\centering
\includegraphics[width=0.3\linewidth]{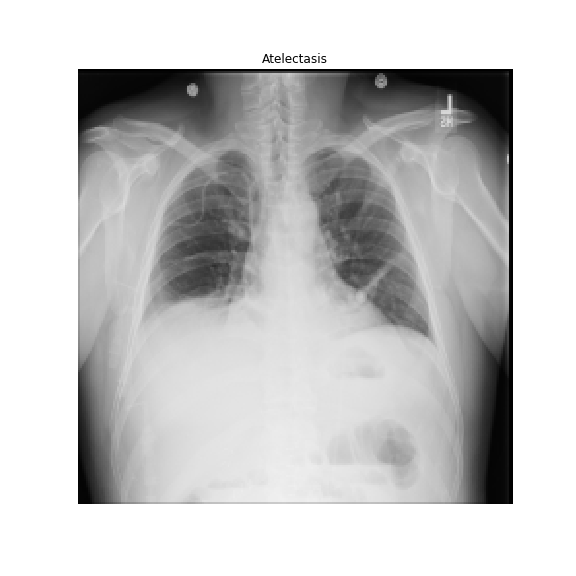}
\includegraphics[width=0.3\linewidth]{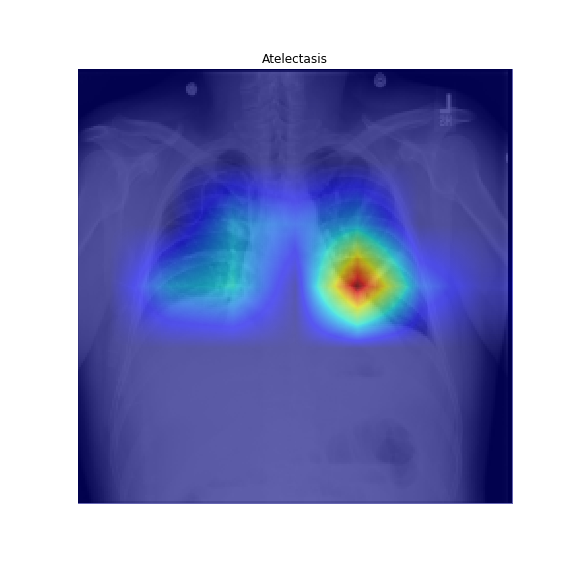}
\includegraphics[width=0.3\linewidth]{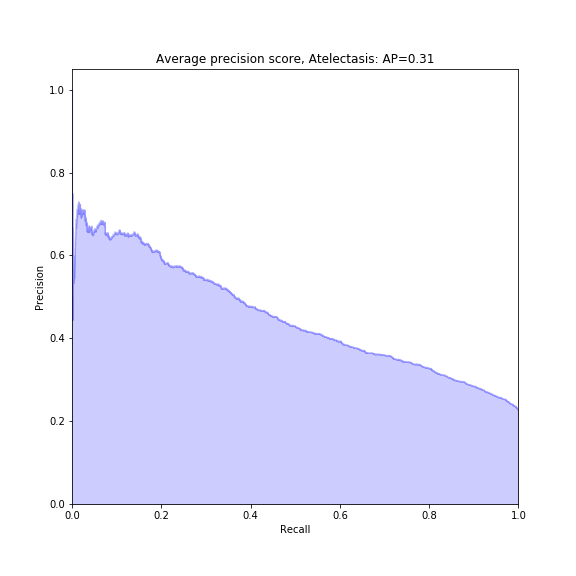}
\caption{Original image (left), Guided Grad-CAM (center) and precision-recall plot (right) for Atelectasis class. The Grad-CAM shows that the CNN is getting very hot spots (regions in red) in the left lung and at the bottom of the right lung. The PR curve for this class gives $AP=0.31$.}
\label{fig15}
\end{figure}

\begin{figure}[h]
\centering
\includegraphics[width=0.3\linewidth]{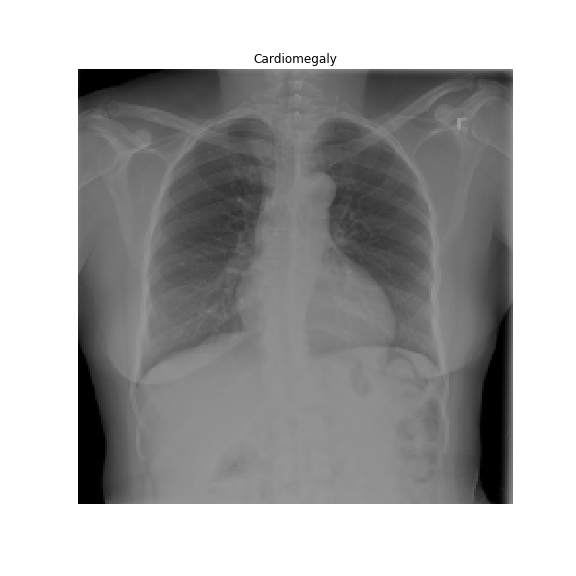}
\includegraphics[width=0.3\linewidth]{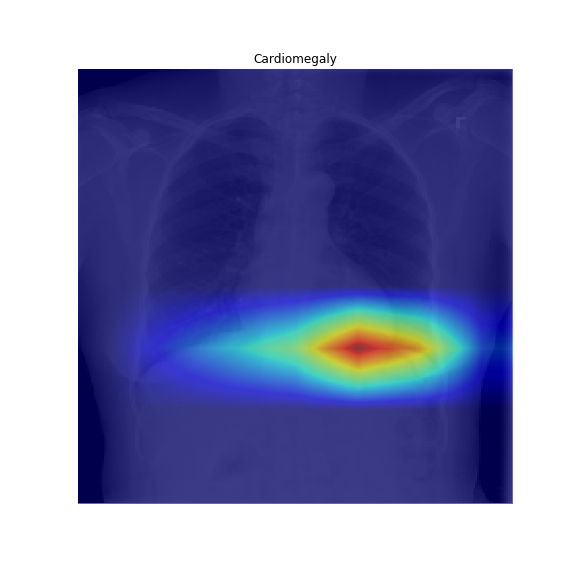}
\includegraphics[width=0.3\linewidth]{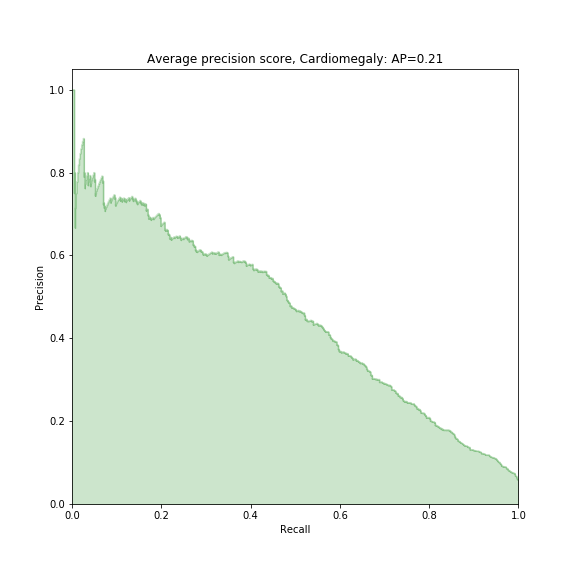}
\caption{Original image (left), Guided Grad-CAM (center) and precision-recall plot (right) for Cardiomegaly class. The Grad-CAM shows that the CNN is getting a very hot spot in the bottom left region. The PR curve for this class gives $AP=0.21$.}
\label{fig17}
\end{figure}

\begin{figure}[h]
\centering
\includegraphics[width=0.3\linewidth]{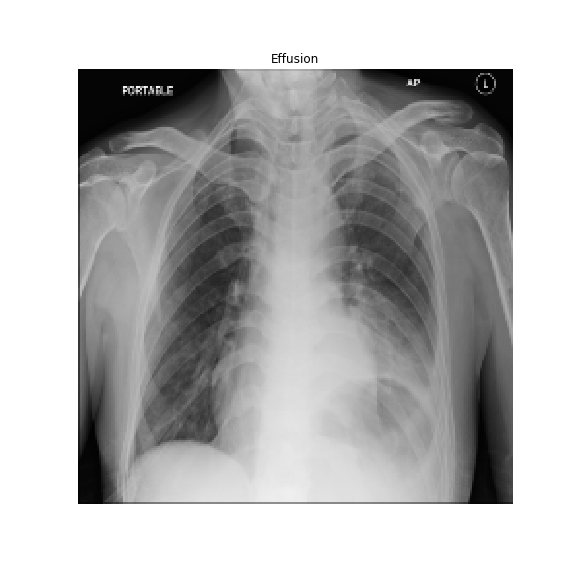}
\includegraphics[width=0.3\linewidth]{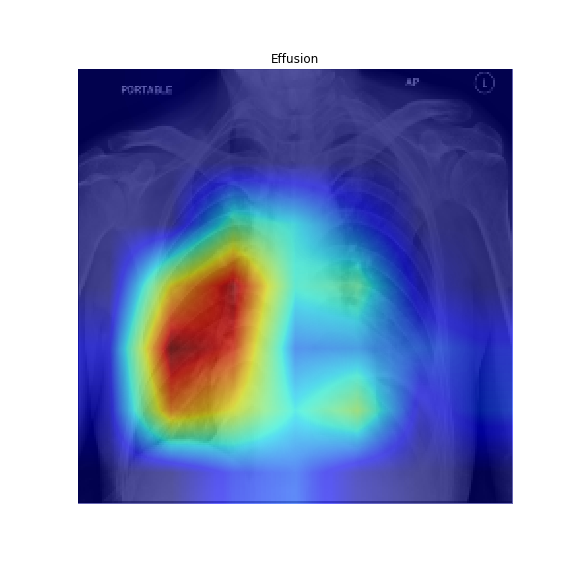}
\includegraphics[width=0.3\linewidth]{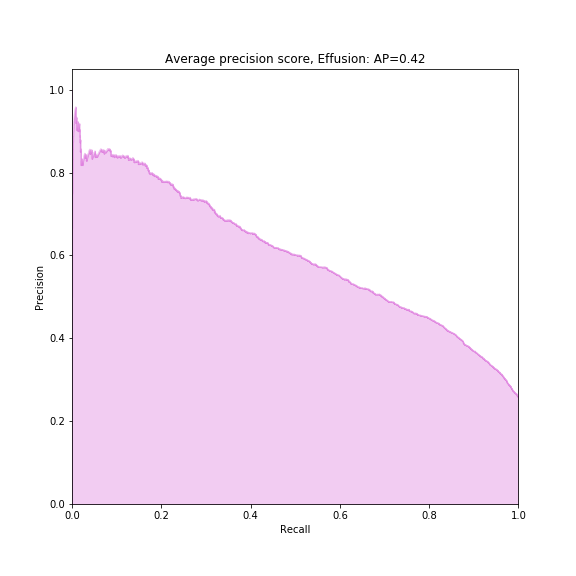}
\caption{Original image (left), Guided Grad-CAM (center) and precision-recall plot (right) for Effusion class. The Grad-CAM shows that the CNN is getting a very hot at the right lung. The PR curve for this class gives $AP=0.42$.}
\label{fig18}
\end{figure}

\begin{figure}[h]
\centering
\includegraphics[width=0.3\linewidth]{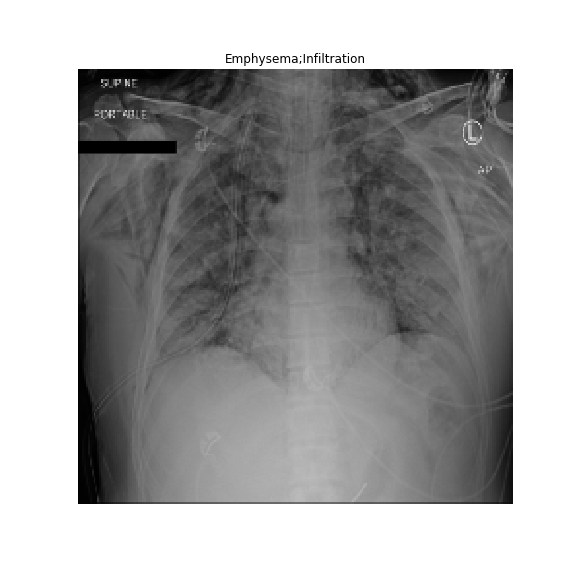}
\includegraphics[width=0.3\linewidth]{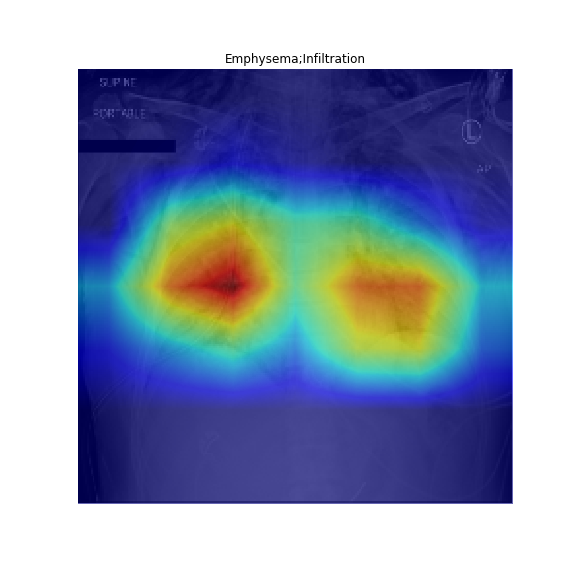}
\includegraphics[width=0.3\linewidth]{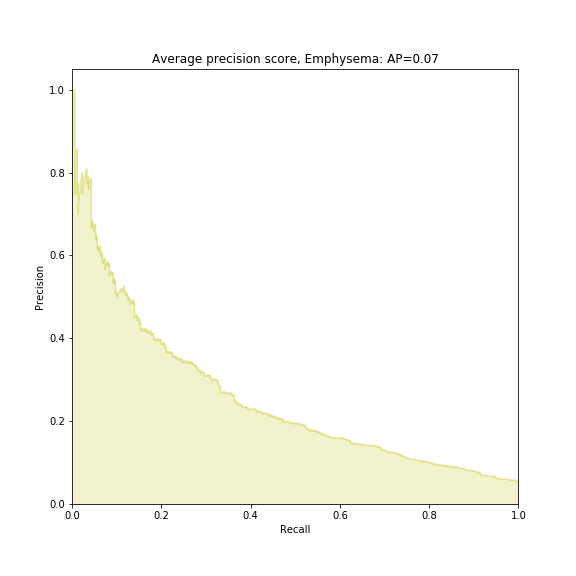}
\caption{Original image (left), Guided Grad-CAM (center) and precision-recall plot (right) for Emphysema class. The Grad-CAM shows that the CNN is getting hot spots at the both lungs. The PR curve for this class gives $AP=0.07$.}
\label{fig19}
\end{figure}

\begin{figure}[h]
\centering
\includegraphics[width=0.3\linewidth]{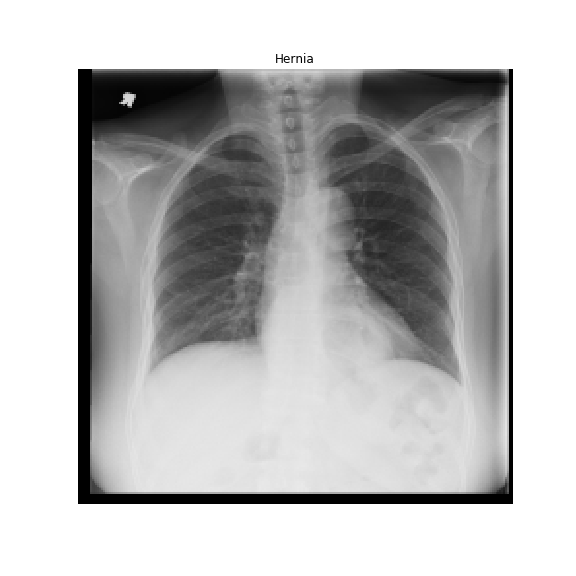}
\includegraphics[width=0.3\linewidth]{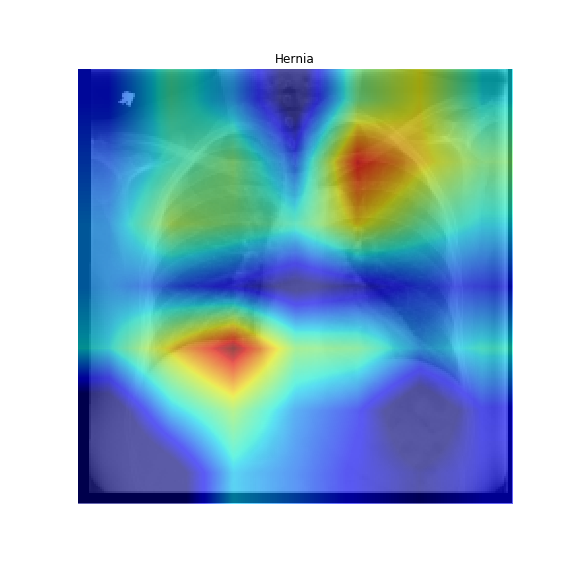}
\includegraphics[width=0.3\linewidth]{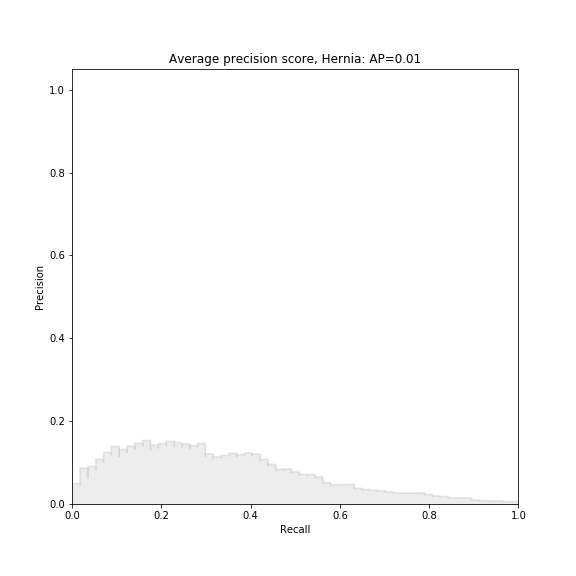}
\caption{Original image (left), Guided Grad-CAM (center) and precision-recall plot (right) for Hernia class. The Grad-CAM shows that the CNN is getting a very hot region on both lungs. The PR curve for this class gives  $AP=0.01$.}
\label{fig21}
\end{figure}

\begin{figure}[h]
\centering
\includegraphics[width=0.3\linewidth]{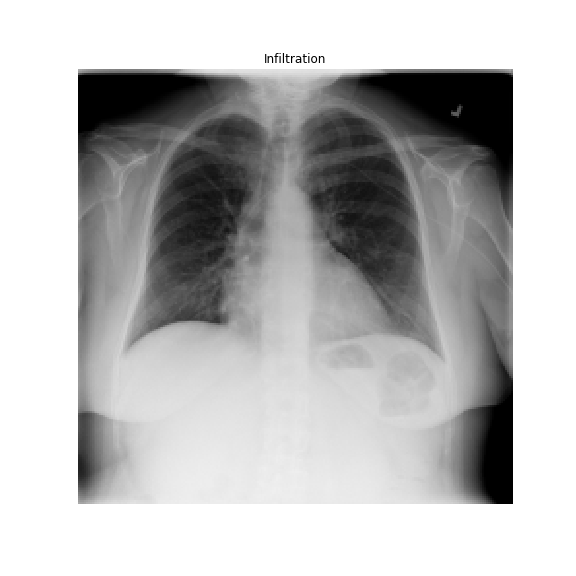}
\includegraphics[width=0.3\linewidth]{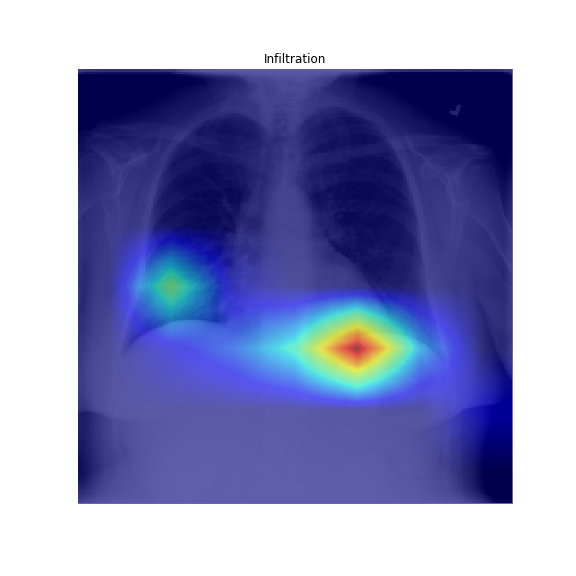}
\includegraphics[width=0.3\linewidth]{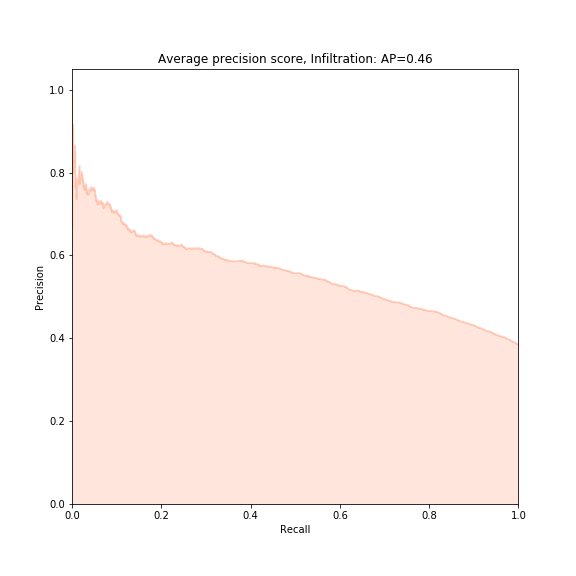}
\caption{Original image (left), Guided Grad-CAM (center) and precision-recall plot (right) for Infiltration class. The Grad-CAM shows that the CNN is getting a very hot region on both lungs, same as the Hernia case. The PR curve for this class gives $AP=0.46$.}
\label{fig22}
\end{figure}

\begin{figure}[h]
\centering
\includegraphics[width=0.3\linewidth]{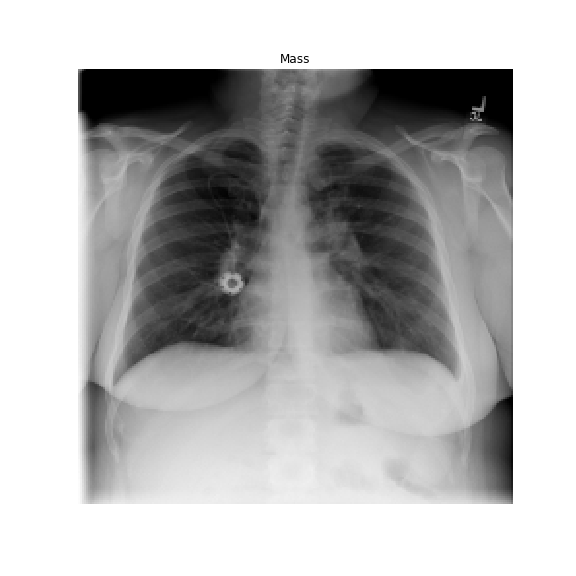}
\includegraphics[width=0.3\linewidth]{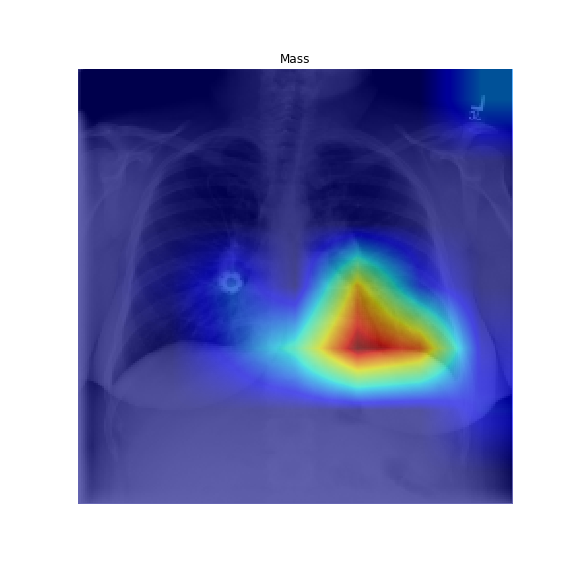}
\includegraphics[width=0.3\linewidth]{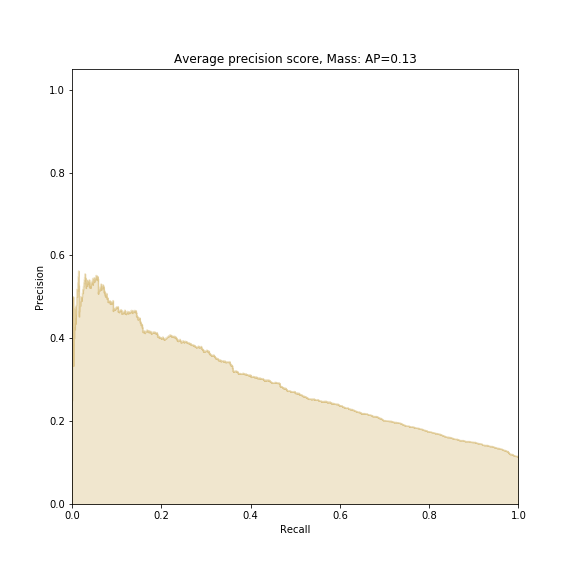}
\caption{original image (left), Guided Grad-CAM (center) and precision-recall plot (right) for Mass class. The Grad-CAM shows that the CNN is getting very hot region on the left lungs. The PR curve for this class gives  $AP=0.13$.}
\label{fig23}
\end{figure}

\begin{figure}[h]
\centering
\includegraphics[width=0.3\linewidth]{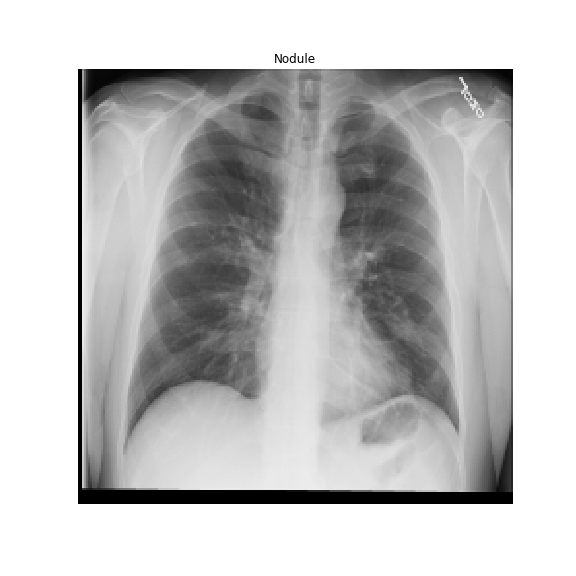}
\includegraphics[width=0.3\linewidth]{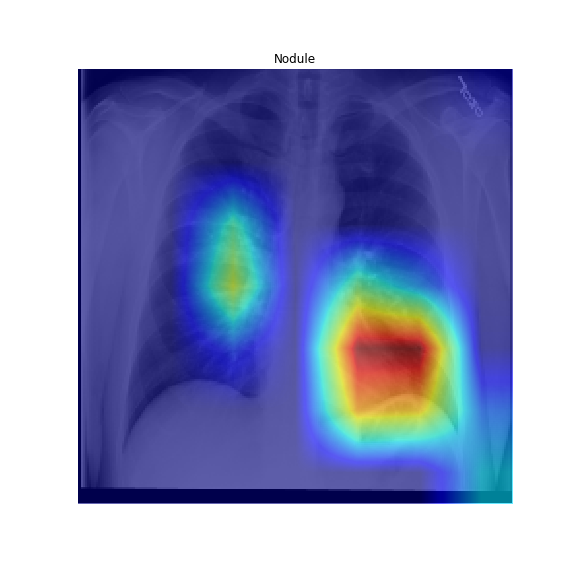}
\includegraphics[width=0.3\linewidth]{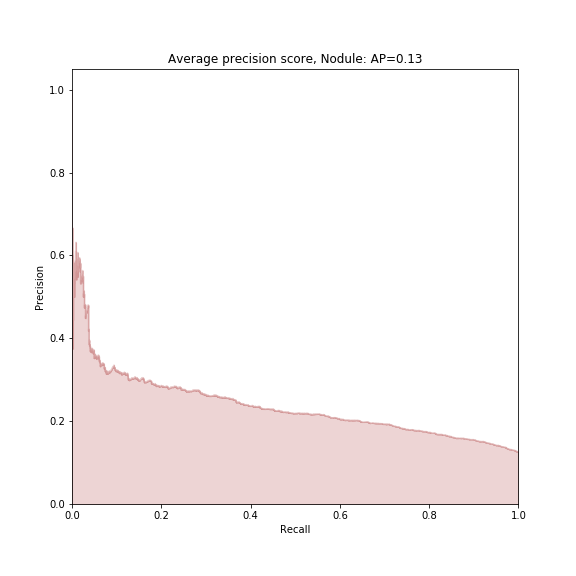}
\caption{Original image (left), Guided Grad-CAM (center) and precision-recall plot (right) for Nodule class. The Grad-CAM shows that the CNN is getting a hot region on left and right lungs. The PR curve for this class gives $AP=0.13$.}
\label{fig24}
\end{figure}

\begin{figure}[h]
\centering
\includegraphics[width=0.3\linewidth]{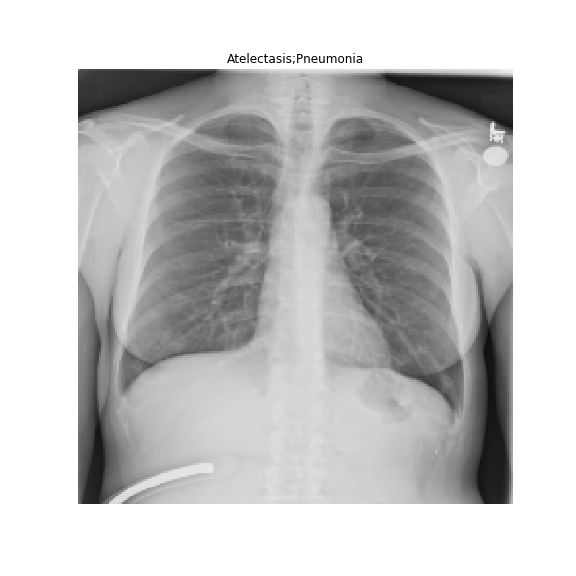}
\includegraphics[width=0.3\linewidth]{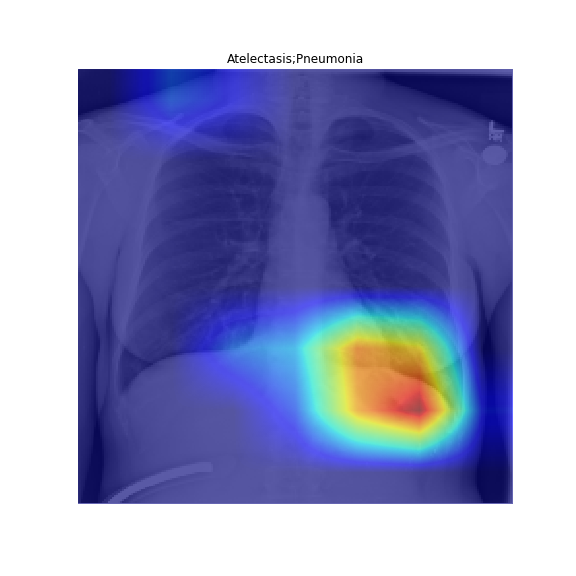}
\includegraphics[width=0.3\linewidth]{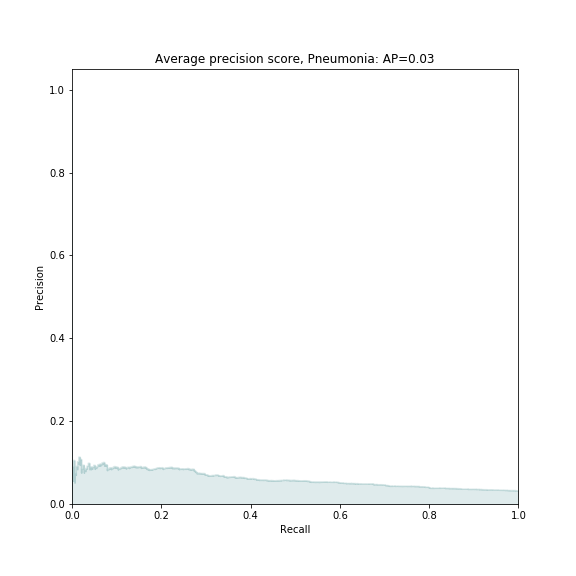}
\caption{original image (left), Guided Grad-CAM (center) and precision-recall plot (right) for Pneumonia class. The Grad-CAM shows that the CNN is getting a hot region on the botom right lung and some warm places all arround lungs and heart. The PR curve for this class gives $AP=0.03$.}
\label{fig26}
\end{figure}

\begin{figure}[h]
\centering
\includegraphics[width=0.3\linewidth]{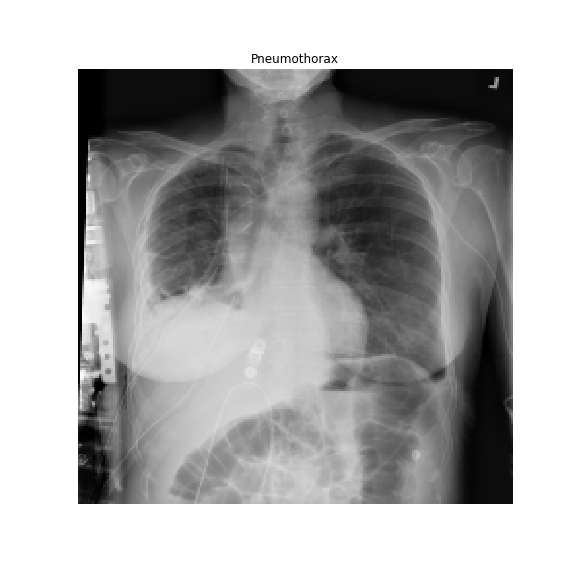}
\includegraphics[width=0.3\linewidth]{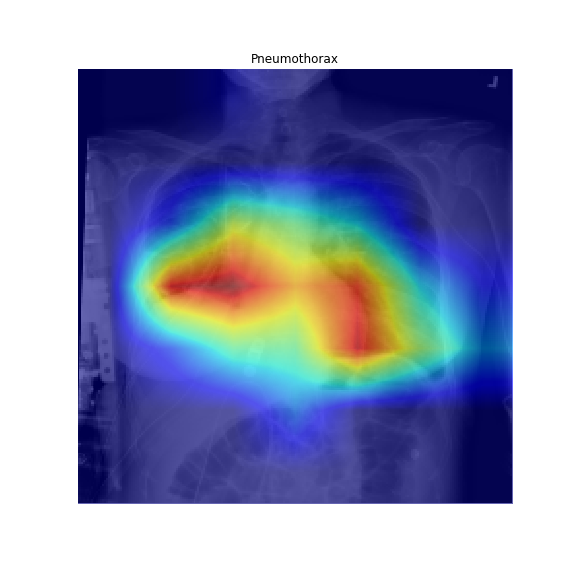}
\includegraphics[width=0.3\linewidth]{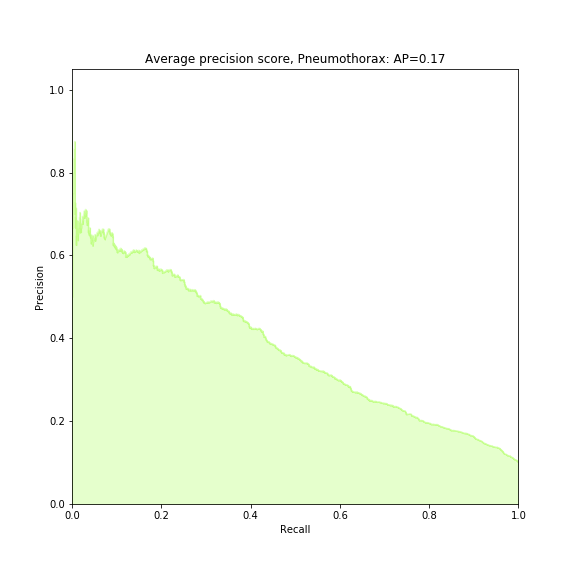}
\caption{Original image (left), Guided Grad-CAM (center) and precision-recall plot (right) for Pneumothorax class. The Grad-CAM shows that the CNN is getting a hot region on both lungs. The PR curve for this class gives $AP=0.17$.}
\label{fig27}
\end{figure}
\clearpage
\subsection{Weighted vs Unweighted (ROC vs PR) for multi-label classification} \label{unweighted}

\begin{figure}[h]
\centering
\includegraphics[width=0.45\linewidth]{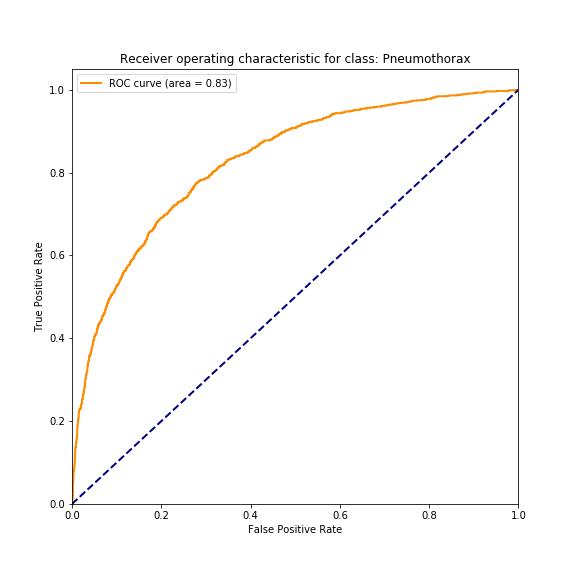}
\includegraphics[width=0.45\linewidth]{figs/AP_plots/Average_precision_score_Pneumothorax.png} 
\vspace{.5cm}
\includegraphics[width=0.45\linewidth]{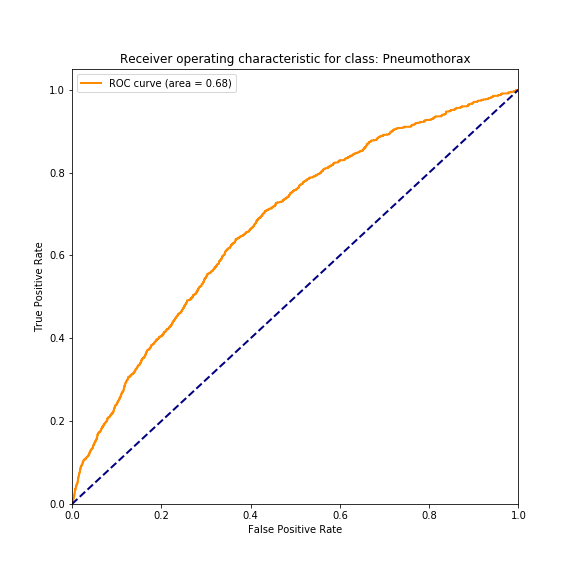}
\includegraphics[width=0.45\linewidth]{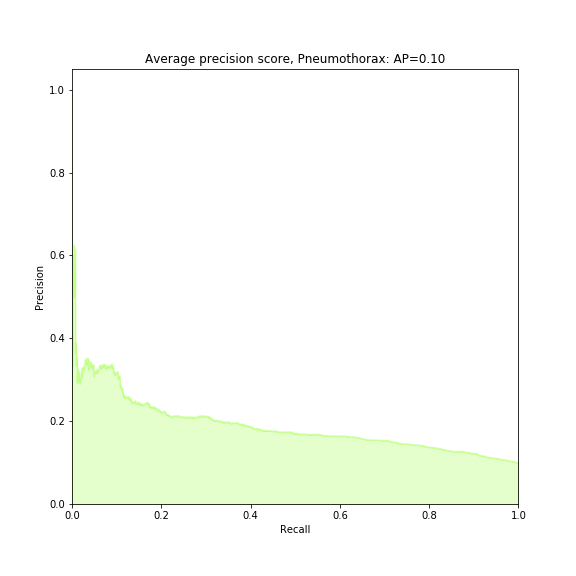}
\caption{{\it Top figures: }PR and ROC for unweighted loss function. The ROC curve is insensitive to changes in the class distribution, while the PR curve can show the effects of imbalance class.
{\it Lower figures: } The ROC curve degrades respect to the previous case (unweighted), whereas the PR shows a more stable behaviour when the recall is increased. The effects of class weights are noticeable in both metrics. }
\label{fig28}
\end{figure}

We noticed that for multi-label classification CheXnet use unweighted binary cross entropy losses. This might not be problematic since in their paper the authors compare the AUC from the ROC metrics with the state of the art models for the ChestX-ray14 dataset. ROC curves have an attractive property: they are insensitive to changes in class distribution. If the proportion of positive and negative classes changes in a test set, the ROC curves will not change. To compare with our results we first implemented our loss function without considering the weights for each class. The results are displayed in the figures \ref{fig10}, \ref{fig11} and \ref{fig12} as well in the tables \ref{tab:comparison}. However, due to the class imbalance the PR curves tell us a very different history from the ROC curves.

Indeed, by comparing Figs.~\ref{fig28} , we see that while the ROC curve for unweighted loss seems to indicate a very good classification, one should consider that reality is different:  we are dealing with a very skewed class distribution, e.g. some labels have samples with order $10^4$ inputs (Atelactasis) while others only a few hundred samples (Hernia) . In this situation, the ROC does not tell the full history of our classification model. We have to also consider cases where we have few samples more carefully.  To use AI techniques in the real world, one should consider how to design and evaluate the performance of such algorithms. If we are dealing with a disease with a small number of examples but still focus on an algorithm which can lead to a precise diagnostic (i.e. high precision) most of the time (i.e. high recall), then we should {\it 1.)} gather more data, which has an intrinsic cost,  and/or {\it 2.) } consider the use of weight class in the loss function, which provides a more realistic measure of accuracy.

\section{Summary and Outlook}

In this note we present our study of the use of Machine Learning techniques to identify diseases using X-ray information. We have used the new framework of \fastai, and imported the resulting model to an app which can be tested by any user. In the app, the user can upload an image and obtain as an output a certain likelihood for pertaining to one of the 14 labelled diseases. This classification could assist diagnosis by medical providers and broaden access to medical services to remote areas.

We have studied two diagnosis situations: first we studied the simpler problem of identifying one disease (pneumothorax) versus any other disease, and then tackled the much more complex case of multiple classification, where we aim to identify individual diseases and the co-occurrence of diseases in the same patient. To emulate realistic situations, we have transformed the high-quality images into lower resolution and applied transformations to them: rotated, cropped and changed the contrast of the images. On those images, we have trained a CNN and applied a number of numerical techniques to speed up computation time, achieving a good accuracy of diagnosis.

We have explored the use of grad-CAM to improve training. We have also studied the different types of measures of how well the algorithm perform. We found that it is particularly important to account for imbalances in both the ROC and Precision-Recall by weighting the samples. For diseases with low numbers of training samples, the issue of weighting may be crucial to get a realistic measure of performance. 

We consider this project as setting the first steps towards a reliable app to help in diagnosing  diseases using images and other sources of information, such as electrocardiogram data. We have provided open source code~\cite{github} for others to use and improve  our procedure as well as a beta version of the app for testing~\cite{heroku} and step-by-step instructions on how to set up the app. We hope this tool can assist clinicians or other health providers in areas where good quality equipment or relevant skills are missing. We welcome any questions and suggestions which can be posted in the GitHub page. 
 
There are a number of improvements in this analysis which one could tackle, including refining the CNN training to improve the diagnosis in the less performing diseases where there is co-occurrence of several diseases, including the no-disease vs disease case, or enlarging the training set with more images from other databases or provided by users, and finally to incorporate more information besides X-ray information.

\clearpage
\appendix

\section{Comparison with CheXnet}

The state-of-the-art in terms of use of the X-ray database~\cite{Wang} is CheXNet~\cite{chexnet}.  In this work, the authors use the same denseNet-121 basic architecture we employed, but in their model the classification layer had one neuron with softmax activation (in our case we have two neurons, one for each class, with logSoftMax activation) and weighted binary cross-entropy loss (in our case we can use a weighted cross-entropy loss). Moreover, we have performed further transformations to the images in the dataset, such as changing the contrast between the dark and bright areas, to emulate more closely difficult conditions a physician could find in a remote location and low-quality equipment. 
 
In the CheXNet they compared with the diagnoses of four radiologists, who studied  a test set with 420 images and labelled them according to the 14 diseases. This was used to evaluate a {\it radiologist F1 score} for the pneumonia detection task and compare to the F1 score obtained by the CheXNet. We had no such benchmarking, and our F1 scores are based on the validation dataset.

Finally, in CheXNet scores were based on unweighted goodness measures which, as we discussed in Sec.~\ref{unweighted}, may not give a realistic view of diseases with small datasets.

\section{Heroku setup}

In this appendix we describe the setup of our online tester for the app. In order to setup the online server we used {\tt docker} and {\tt heroku}. After we have trained our model, we export it using the \fastai function \verb|export|, which exports both the model and weights. The model is saved in a pkl (pickle) format and the weights are saved in pth (pytorch) format, which we store in a folder to upload to the {\tt heroku} webserver.

\paragraph{Docker:}

We recommend to install {\tt docker}, which facilitates the sharing of codes between any machines, without the need to install many dependencies. The installation of {\tt docker} is quite straightforward~\url{https://docs.docker.com/install/#support}, except when running on linux where you might need to take some extra steps \url{https://docs.docker.com/install/linux/linux-postinstall/} to setup the proxies properly.

After the installation, one can create a docker image for our model to work and {\tt docker} allows us to create a virtual machine able to run in any kind of computer as long it has the docker tool.  To create the image one first needs to create a empty file called \textit{Dockerfile}, the \textit{Dockerfile} will contain the instructions needed to load the model and run it in our environment.
In the following we give some examples of how to do this:

\begin{verbatim}
FROM python:3.6-slim-stretch

RUN apt update && \
    apt install -y python3-dev gcc

WORKDIR app

ADD requirements.txt .
RUN pip install -r requirements.txt
#pip install --no-cache-dir -r
ADD models models
ADD src src

EXPOSE 80

# Run it once to trigger DenseNet download
RUN python src/app.py prepare

# Start the server
CMD ["python", "src/app.py", "serve"]

\end{verbatim}

To call the  pre-installed docker image with python3 and gcc compiler from the docker website we do the following:
\begin{verbatim}
FROM python:3.6-slim-stretch

RUN apt update && \
    apt install -y python3-dev gcc
\end{verbatim}

\begin{verbatim}
ADD requirements.txt .
RUN pip install -r requirements.txt
\end{verbatim}
here we add the file requirements to docker so it download and install the libraries we are using, this files contains the following instructions:
\begin{verbatim}
torch==1.0.0
torchvision==0.2.1
Flask==1.0.2
\fastai==1.0.50.post1
\end{verbatim}
these are the packages and libraries with their respective versions. The following line will instruct pip to download and install the packages in the Docker image we want to create.
\begin{verbatim}
ADD models models
ADD src src
\end{verbatim}
which tells docker the paths of our source files, where we will put the app program and other files we will need to run the app, and the path to our model and weights.

The line:
\begin{verbatim}
EXPOSE 80
\end{verbatim}
tells docker which proxy port our app will run, important to our app.

And the following lines tell docker to run our app stored in the folder \textit{src}:
\begin{verbatim}
# Run it once to trigger DenseNet download
RUN python src/app.py prepare

# Start the server
CMD ["python", "src/app.py", "serve"]

\end{verbatim}

\paragraph{The app:}
We wrote the app in a python script file called app.py. The app is written using the flask library which can create web interfaces based in java and php. The main important  features of the web app are the \verb|config.yaml| file and the \textit{static} folder. The \verb|config.yaml| file defines the url configurations of the web app for the example images and other links, while the folder \textit{static} contains the css scripts for colours, button sizes, etc.

After we prepare the app and setup the docker file, we now can create the docker image by executing the command:
\begin{verbatim}
docker build --tag=ml-xray-v-0-1 .
\end{verbatim}
where the \verb|tag| flag is to indicate the name of the image we create. Now we can start to set up the Heroku webserver.

\paragraph{Heroku:}
First, one  has to create a Heroku account and then  install the heroku git\footnote{You might need to install git, please follow these instructions: \url{https://git-scm.com/book/en/v2/Getting-Started-Installing-Git}.} app in your machine. 

Next, open a terminal and run the command to login into your heroku account:
\begin{verbatim}
heroku login
\end{verbatim}
which will ask for your login name, usually your email, and password. After login, one has to pull the folder which contains our dockerfile, model and source code of our app:
\begin{verbatim}
heroku git:remote -a ml-xray
\end{verbatim}
Now we can give the instruction to heroku to push our docker container, by running the command:
\begin{verbatim}
heroku container:login
heroku container:push web --app ${ml-xray}
\end{verbatim}
and release the container in the webserver:
\begin{verbatim}
heroku container:release web --app ${ml-xray}
\end{verbatim}
Finally, one can now open the app and get some logs by running the command:
\begin{verbatim}
heroku open --app $ml-xray
heroku logs --tail --app ${ml-xray}
\end{verbatim}
Note that after 15 minutes of inactivity the server goes to sleep mode and it might take a little while to wake up again.


\end{document}